\documentclass[10pt,twocolumn,letterpaper]{article}

\usepackage[accsupp]{axessibility}  
\usepackage{cvpr}            
\usepackage{amsfonts}
\usepackage{amsmath}
\usepackage{amsthm}
\usepackage{bm}
\usepackage{multirow}
\usepackage{multicol}
\usepackage{graphicx}
\usepackage{enumitem}
\usepackage{array}
\usepackage{wrapfig}
\usepackage{xcolor}
\usepackage{shadethm}
\usepackage{thmtools}
\usepackage{algorithm}
\usepackage{algpseudocode}
\usepackage{nicefrac}
\usepackage{microtype}
\usepackage{booktabs}
\usepackage[para,online,flushleft]{threeparttable}
\usepackage{pifont}
\usepackage[labelfont=bf,textfont=it]{caption}
\usepackage{textcomp, gensymb}

\declaretheoremstyle[%
  spaceabove=-6pt,%
  spacebelow=6pt,%
  headfont=\bfseries\itshape,%
  postheadspace=0.5em,%
  qed=\qedsymbol%
]{mystyle}

\theoremstyle{mystyle}

\definecolor{cvprblue}{rgb}{0.21,0.49,0.74}
\usepackage[pagebackref,breaklinks,colorlinks,allcolors=cvprblue]{hyperref}
\newcommand{\colref}[3]{\hyperref[#2]{#1~\ref*{#2}{#3}}}
\newcommand{\figref}[1]{\colref{Figure}{#1}{}}

\newcommand{\tabref}[1]{\colref{Table}{#1}{}}

\newcommand{\algoref}[1]{\colref{Algorithm}{#1}{}}

\title{SC-NeRF: NeRF-based Point Cloud Reconstruction using a Stationary Camera for Agricultural Applications}

\author{Kibon Ku,
Talukder Z Jubery,
Elijah Rodriguez,
Aditya Balu,
Soumik Sarkar,\\
Adarsh Krishnamurthy$^*$,
Baskar Ganapathysubramanian$^*$\vspace{0.5em}\\
Iowa State University, Ames, IA, USA.\\
{\small emails: \texttt{kibona9|znjubery|eli320|baditya|soumiks|adarsh|baskarg@iastate.edu}}}

\begin{document}
\twocolumn[{
\renewcommand\twocolumn[1][]{#1}
    \maketitle
    \centering
    \includegraphics[width=0.4\linewidth]{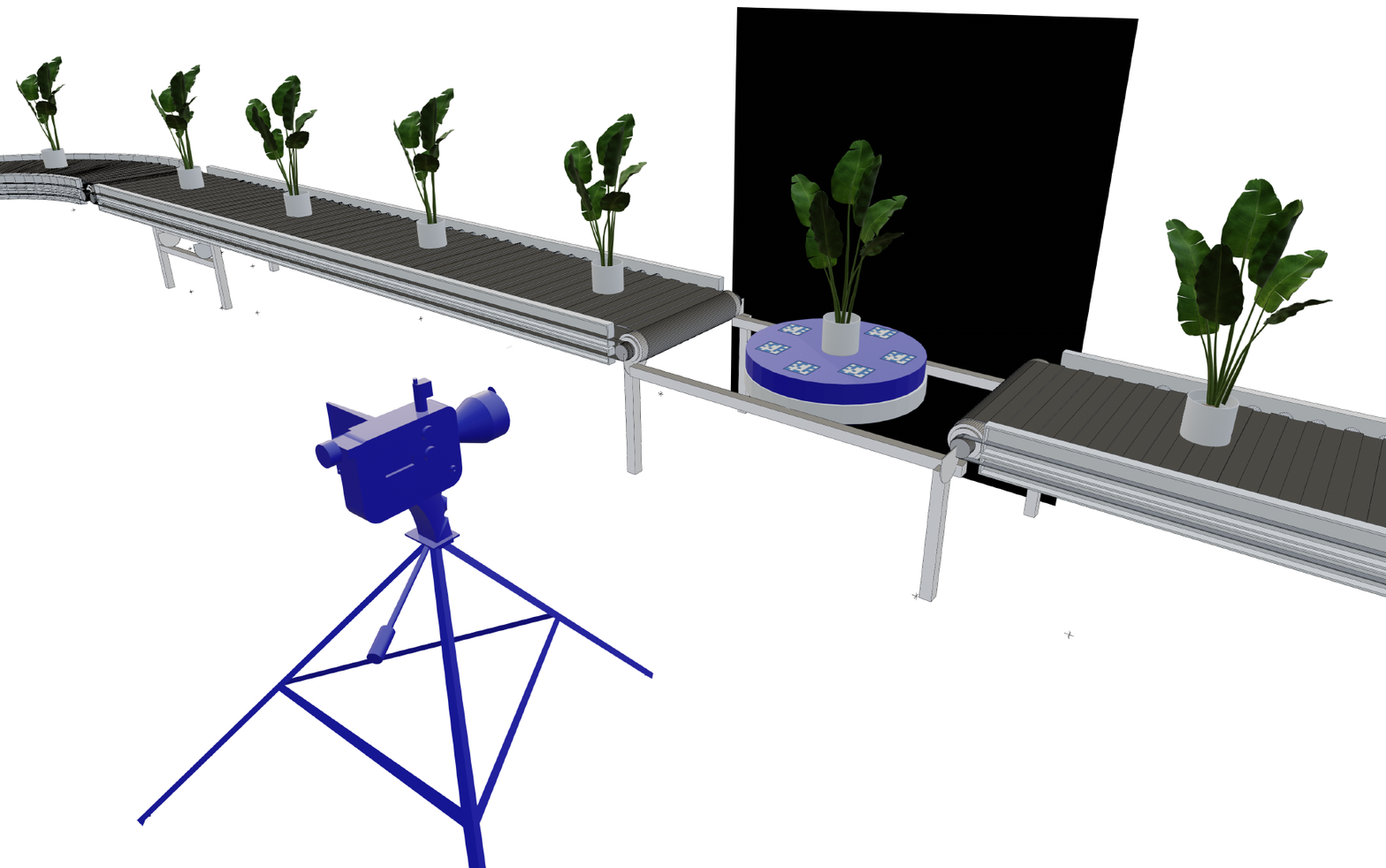}
    \includegraphics[width=0.58\linewidth]{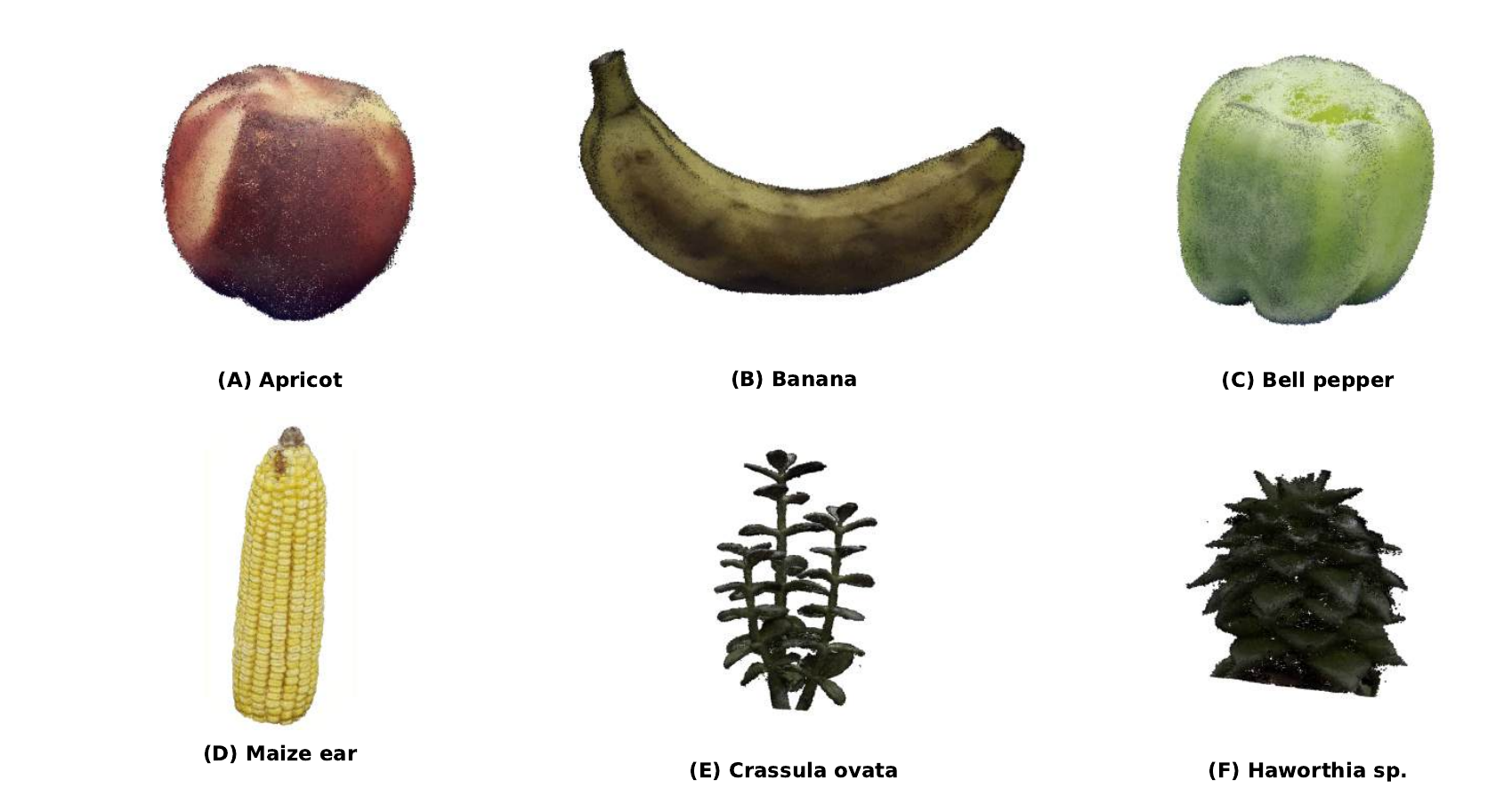}
    \captionof{figure}{Schematic of the stationary camera imaging system for NeRF-based point cloud reconstruction in high-throughput plant phenotyping. In this setup, each plant is conveyed to a rotating turntable marked against a matte black background. Over a full 30-second rotation, a tripod-mounted stationary camera captures high-resolution images that serve as input for NeRF techniques to generate 3D reconstructions. This streamlined approach eliminates the need for complex moving-camera rigs, aligning with the objectives of efficient, scalable agricultural imaging. The right shows different PCD reconstruction using the stationary camera. (a) Apricot, (b) Banana, (c) Bell pepper, (d) Maize ear, (e) \emph{Crassula ovata}, and (f) \emph{Haworthia sp.}\vspace{1em}}
    \label{fig:teaser}
}]

\begin{abstract}
This paper presents a NeRF-based framework for point cloud (PCD) reconstruction, specifically designed for indoor high-throughput plant phenotyping facilities. Traditional NeRF-based reconstruction methods require cameras to move around stationary objects, but this approach is impractical for high-throughput environments where objects are rapidly imaged while moving on conveyors or rotating pedestals. To address this limitation, we develop a variant of NeRF-based PCD reconstruction that uses a single stationary camera to capture images as the object rotates on a pedestal. Our workflow comprises COLMAP-based pose estimation, a straightforward pose transformation to simulate camera movement, and subsequent standard NeRF training. A defined Region of Interest (ROI) excludes irrelevant scene data, enabling the generation of high-resolution point clouds (10M points). Experimental results demonstrate excellent reconstruction fidelity, with precision-recall analyses yielding an F-score close to 100.00 across all evaluated plant objects. Although pose estimation remains computationally intensive with a stationary camera setup, overall training and reconstruction times are competitive, validating the method's feasibility for practical high-throughput indoor phenotyping applications. Our findings indicate that high-quality NeRF-based 3D reconstructions are achievable using a stationary camera, eliminating the need for complex camera motion or costly imaging equipment. This approach is especially beneficial when employing expensive and delicate instruments, such as hyperspectral cameras, for 3D plant phenotyping. Future work will focus on optimizing pose estimation techniques and further streamlining the methodology to facilitate seamless integration into automated, high-throughput 3D phenotyping pipelines. We provide all datasets and our code, available at  \url{https://baskargroup.github.io/SC-NeRF/}
\end{abstract}

\pagebreak

\section{Introduction}
\label{sec:intro}


Accurate characterization of plant phenotypes is crucial for improving crop yield, resilience, and sustainability in agriculture~\citep{kusmec2018harnessing, blancon2024maize}. Advanced 3D phenotyping techniques enable precise measurement of critical traits, including plant architecture, leaf angles, and biomass allocation, significantly impacting yield prediction and environmental adaptability~\citep{grys2017machine, westhues2021prediction, gupta2024ai, tucker2020evaluating}. Given the growing global need for sustainable agriculture, robust and scalable 3D phenotyping methods are indispensable for advancing crop improvement and breeding programs.

Conventional approaches to 3D phenotyping primarily involve photogrammetry techniques such as structure-from-motion (SfM) and multi-view stereo (MVS)\citep{eltner2020structure, chen2019point}, as well as terrestrial laser scanning (TLS)\citep{medic2023challenges,young2023canopy}. Although these methods provide detailed structural data and have been effectively applied across various crops~\citep{hu2020nondestructive, lei20233d}, they present several practical limitations, including high equipment costs, manual labor, and significant computational demands~\citep{andujar2018three, tang2022benefits}. Additionally, their scalability and capacity to capture minute structural details in dynamic agricultural scenarios are limited~\citep{paturkar2021effect, lu2023bird}.

Recent advancements in artificial intelligence (AI), particularly Neural Radiance Fields (NeRF), have opened new avenues for detailed and scalable 3D reconstruction~\citep{mildenhall2021nerf}. NeRF utilizes deep learning to implicitly represent volumetric scenes, synthesizing photorealistic views from multiple 2D images without explicit geometric constraints. Its resolution-invariant representation offers advantages in capturing intricate plant features compared to traditional methods~\citep{feng20233d, cuevas2020segmentation, sarkar2023cyber}. AI-based NeRF approaches thus present significant potential for rapid, cost-effective, and accurate 3D plant phenotyping.


\begin{figure*}[ht!] 
    \centering
    \includegraphics[width=0.99\linewidth]{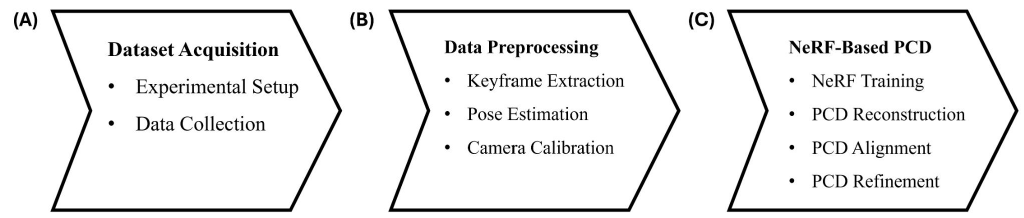}
    \caption{Workflow of the NeRF-based 3D reconstruction pipeline. The process consists of three main steps: (A) Dataset Acquisition, where the experimental environment is set up, and multi-view image data is collected using a stationary camera; (B) Data Preprocessing, involving Keyframe extracion, pose estimation and camera calibration to ensure geometric consistency; and (C) NeRF-Based PCD, where a NeRF model is trained for scene representation, followed by PCD Reconstruction, Alignment, and Refinement to generate high-quality 3D point clouds. This structured approach improves the accuracy and scalability of 3D reconstruction for phenotyping and other agricultural vision applications.}
    \label{fig:figure2}
\end{figure*}
However, conventional NeRF implementations require cameras to move around stationary objects to capture multiple viewpoints, presenting significant logistical challenges in high-throughput indoor phenotyping environments. High-throughput indoor phenotyping has become increasingly widespread and is now a standard method to rapidly evaluate plants and agricultural products in breeding, plant science, and agricultural production applications. Facilities routinely rely on automated conveyors and rotating pedestals to quickly image large numbers of plants, ensuring efficient data collection, consistency, and operational throughput. In this context, a limitation of current high-throughput 3D NeRF based phenotyping methods is their dependence on manual or mechanically-assisted camera movement. Most phenotyping systems currently in use require either manual camera rotation or mechanical platforms to capture images from multiple viewpoints, introducing operational inefficiencies. These approaches increase labor and operational costs and restrict data collection frequency, throughput, and scalability, particularly in high-throughput scenarios common in agricultural breeding and research facilities. In such environments, stationary camera setups are essential for maintaining workflow efficiency, uniformity, and reproducibility. Consequently, traditional moving-camera NeRF approaches are impractical for these standardized indoor phenotyping scenarios, highlighting the need for alternative methods specifically tailored to stationary imaging setups.

Moreover, other high-end imaging solutions, such as LiDAR scanners and multi-camera setups, involve prohibitive upfront investments and high maintenance costs, making them impractical for widespread adoption. Frequent recalibration, complex operational requirements, and limited capacity for continuous monitoring further reduce their suitability, especially in indoor phenotyping facilities where rapid and consistent data acquisition is critical. Additionally, these imaging techniques often encounter challenges related to data consistency and quality. Variations in lighting conditions, occlusions caused by complex plant structures, and motion blur resulting from mechanical movements degrade the quality and reliability of the resulting 3D reconstructions. Overcoming these challenges by eliminating the reliance on camera movement while maintaining high-quality and consistent data acquisition would significantly enhance the efficiency and effectiveness of agricultural 3D phenotyping systems.

To address these limitations, this paper presents a stationary-camera-based NeRF framework explicitly developed for indoor high-throughput phenotyping. Unlike moving-camera setups, our approach uses a rotating pedestal, cutting cost and complexity while supporting sensitive tools like hyperspectral cameras. Our approach integrates COLMAP-based pose estimation and a simple pose transformation to simulate camera movement, enabling standard NeRF training using stationary-camera data. We demonstrate the effectiveness and scalability of our method, achieving high-fidelity point cloud reconstructions with near-perfect precision-recall metrics. Our primary contributions include: (1) a novel stationary-camera NeRF reconstruction pipeline designed specifically for high-throughput indoor phenotyping, (2) extensive experimental validation demonstrating reconstruction fidelity, and (3) evidence of computational feasibility, paving the way for seamless integration into automated phenotyping workflows.

\section{Background}

Recent advances in neural implicit representations have significantly improved 3D reconstruction from 2D images across various domains. In agriculture, NeRF-based methods have been explored using diverse camera setups \cite{arshad2024evaluating,hu2024high,wu163d,jignasu2023plant}. For example, Hu et al. \cite{hu2024high} demonstrated high-fidelity reconstructions of plants in both indoor and outdoor orchards with a moving camera, while Wu et al. \cite{wu163d} utilized a rotating camera rig to capture multi-angle videos in indoor settings. Gao et al. \cite{gao2023mc} further contributed by employing a fixed multi-camera system for reconstructing indoor objects under controlled conditions.

Despite these achievements, fixed, stationary cameras remain relatively underexplored for NeRF-based reconstruction. Traditional photogrammetric techniques—such as voxel carving from silhouettes captured by fixed cameras—offer a foundation \cite{hirano20093d,zhang2008self,feng20233d}, but standard NeRF approaches struggle with varying illumination in static scenes. To address these challenges, several recent works have proposed alternative strategies. For instance, EventNeRF \cite{rudnev2023eventnerf} leverages event-based cameras to enhance reconstruction under rapid motion and low-light conditions. SII-NeRF Scans employs structured illumination to achieve high-quality results, although its reliance on a large, controlled scanning environment limits its portability. Additionally, research on unposed turntable images has shown promise in reducing the dependency on computationally intensive pose estimations \cite{levy2023melon}, thereby streamlining data acquisition. Moreover, Hyperspectral Neural Radiance Fields \cite{chen2024hyperspectral} introduces a stationary hyperspectral camera system that captures rich geometric, radiometric, and spectral details, a capability especially valuable for applications demanding precise color and spectral resolution.

Building on these advances, our work proposes a three-channel approach tailored for agricultural applications. Our goal is to develop a method that is robust, low-cost, and high-throughput by optimizing image resolution, reducing the number of required images, and ensuring accurate color representation. This approach minimizes setup constraints, such as those required by OB-NeRF \cite{wu163d}, while delivering high-fidelity 3D reconstructions, making it ideally suited for real-world agricultural scenarios.

\section{Methodology}
\label{sec:Workflow}

\figref{fig:figure2} lays out our workflow. We describe each step in detail below.

\begin{figure*}[h] 
    \centering
    \includegraphics[width=0.85\linewidth]{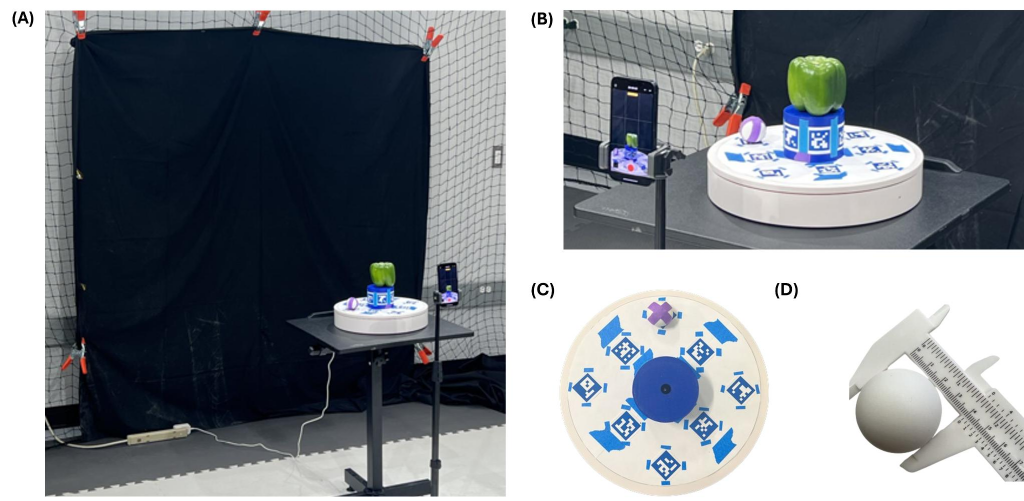}
    \caption{Experimental setup. (A) Overall setup, where a stationary camera (iPhone 13 Mini) records a rotating object (green bell pepper) placed on a turntable against a black matte fabric to minimize background noise and improve segmentation. (B) Close-up of the turntable and object, highlighting the elevated platform and ArUco markers used for pose estimation and structured scene reconstruction. (C) ArUco markers for pose estimation, where different types of markers are used for feature matching in COLMAP to compute camera poses. (D) Scale calibration, where a ping pong ball (radius = 0.04 m) is measured with a caliper to ensure accurate scaling in the reconstructed point cloud data (PCD). This setup enables precise alignment between the stationary camera’s PCD measurements and the rotating camera’s ground-truth data for quantitative evaluation.}
    \label{fig:Exp_Setup}
\end{figure*}

\subsection{Experimental Setup}
The experimental setup (\figref{fig:Exp_Setup}) was designed to capture video data using an iPhone 13 mini under two different conditions: one where the object was placed on a rotating turntable while the camera remained stationary and another (our baseline, standard NeRF approach) where the object remained fixed while the camera moved around it to capture different perspectives. The latter served as the ground truth for evaluating the quality of PCD reconstruction.

To ensure stability and consistency, the iPhone 13 mini was mounted on a tripod and positioned orthogonally to the object. As shown in \figref{fig:Exp_Setup}, the camera screen was adjusted to frame the object and ArUco markers, ensuring proper alignment while minimizing unwanted elements. A black matte fabric background was used to enhance object segmentation and eliminate visual distractions. The setup was placed on a vibration-free surface with adequate clearance between the object and the background to maintain uniform imaging conditions.

For pose estimation and structured scene reconstruction, 5×5 ArUco markers were generated using the ArUco Markers Generator (https://chev.me/arucogen/). The markers were positioned in two configurations: six markers were attached to a 3D-printed blue cylinder (diameter = 0.09 m, height = 0.07 m), which was digitally designed and 3D-printed using an FDM-based desktop printer. A custom cylindrical container was designed to enhance visual clarity and facilitate reliable marker tracking during reconstruction. The container featured a uniformly colored surface to minimize artifacts caused by blending with the black background. Its constant radius ensured smooth and continuous visibility of the attached markers, addressing the geometric inconsistencies of standard plant pots. In addition, eight markers were placed on a circular paper mounted to the turntable to maximize pose estimation support. This was particularly important for the stationary camera setup, where only the object and markers change between frames—unlike rotating camera environments, which provide natural parallax and viewpoint diversity. These markers provided consistent feature points for COLMAP, enabling accurate computation of camera extrinsic parameters and contributing to robust structured scene reconstruction. A ping pong ball (radius = 0.04 m) was included in the scene for metric scale calibration, measured using a digital caliper to ensure accurate dimension scaling.

\subsection{Data Acquisition}
 The dataset included six objects of varying shapes, and geometric complexities: apricot, paprika, banana, maize ear (corn cob), and two potted plants, \textit{Haworthia sp.}, and \textit{Crassula ovata}. Each object was placed individually on a motorized turntable rotating at a constant speed to ensure uniform coverage. Each video was recorded for 30 seconds at 4K resolution (3840×2160) at 30 fps using HEVC (Main 10, BT.2020 color space) encoding, preserving high dynamic range fidelity. In the stationary camera with rotating object configuration, the turntable maintained a constant speed to ensure complete object coverage. In the stationary object with moving camera configuration (our baseline comparison), the camera was manually moved around the object to capture multiple viewpoints. Each imaging protocol was repeated three times, and the highest-quality recording was selected for analysis.

This controlled experimental design ensured alignment between stationary-camera PCDs and ground-truth PCDs generated from the moving-camera recordings. The integration of ArUco markers for pose estimation and a known-scale reference object provided a reproducible and structured framework for evaluating PCD reconstruction accuracy.


\subsection{Data Preprocessing}

The data preprocessing pipeline consisted of (1) keyframe extraction, (2) pose estimation, and (3) camera calibration to ensure accurate point cloud reconstruction.

The recorded video was converted into frames at 4 frames per second (FPS) using \texttt{FFmpeg} to balance computational efficiency and feature tracking accuracy. The extracted images were stored in a structured dataset directory for subsequent processing. The optimal frame rate for capturing slow, structured motion, such as a rotating camera, is typically 4–5 FPS, as it captures sufficient detail while minimizing redundancy~\cite{delattre2023robust}. This frame rate is particularly effective for scenarios involving predictable and gradual motion, such as object or camera rotations, where smooth motion can be maintained without requiring excessive frame rates that introduce unnecessary data overhead. 

Motivated by prior findings that 4–5 FPS is typically sufficient for structured motion~\cite{delattre2023robust}, we adopted a top-down strategy for frame rate selection in ~\algoref{alg:select-fps}. Starting from 5 FPS, we progressively evaluated lower frame rates to determine the minimal rate that still achieved complete image registration in COLMAP. For each candidate FPS, we measured the number of registered images and selected the lowest FPS that yielded 100\% registration while minimizing the total number of extracted frames. This approach avoids unnecessary redundancy and ensures data efficiency without compromising reconstruction quality.

\begin{algorithm}[t!]
\caption{\textsc{SelectOptimalFPS}: Find Minimum FPS with 100\% Registration}
\label{alg:select-fps}
\begin{algorithmic}
\Require Raw video file $\mathcal{V}$, candidate frame rates $\texttt{fps\_list}$
\Ensure Optimal frame rate $\texttt{fps}_{\text{opt}}$

\State Initialize: $\texttt{fps}_{\text{opt}} \gets \texttt{None}$, $\texttt{min\_frames} \gets \infty$

\For{each $\texttt{fps}$ in $\texttt{fps\_list}$}
    \State Extract frames at $\texttt{fps}$ using \texttt{ffmpeg}
    \State Run COLMAP: feature extraction, matching, SfM
    \State Count registered images: $\texttt{N}_{\text{reg}}$
    \State Count total extracted frames: $\texttt{N}_{\text{frames}}$
    \If{$\texttt{N}_{\text{reg}} = \texttt{N}_{\text{frames}}$ \textbf{and} $\texttt{N}_{\text{frames}} < \texttt{min\_frames}$}
        \State $\texttt{fps}_{\text{opt}} \gets \texttt{fps}$
        \State $\texttt{min\_frames} \gets \texttt{N}_{\text{frames}}$
    \EndIf
\EndFor

\Return $\texttt{fps}_{\text{opt}}$
\end{algorithmic}
\end{algorithm}

\begin{algorithm}[t!]
\caption{Custom Pose Estimation and Preprocessing Pipeline}
\label{alg:pose-estimation}
\begin{algorithmic}[1]
\Require Raw video file $\mathcal{V}$ in \texttt{.MOV} format
\Ensure Camera-to-world pose matrices $\{ \mathbf{T}_{wc}^{(i)} \}$ and processed image data for NeRF

\State \textbf{Frame Extraction:}
\Statex \hspace{1em} $\texttt{fps}_{\text{opt}} \gets \textsc{SelectOptimalFPS}(\mathcal{V})$
\Statex \hspace{1em} Extract frames $\{ \mathbf{I}_i \}$ from $\mathcal{V}$ using $\texttt{fps}_{\text{opt}}$ via \texttt{ffmpeg}

\State \textbf{COLMAP Database Initialization:}
\Statex \hspace{1em} Create empty database for feature storage

\State \textbf{Feature Extraction:}
\Statex \hspace{1em} Use SIFT with GPU acceleration to extract local features from each $\mathbf{I}_i$

\State \textbf{Feature Matching:}
\Statex \hspace{1em} Perform sequential matching to match features between consecutive frames

\State \textbf{Structure-from-Motion (SfM):}
\Statex \hspace{1em} Run COLMAP's \texttt{mapper} to:
\Statex \hspace{2em} Estimate camera poses $\{ \mathbf{T}_{cw}^{(i)} \}$
\Statex \hspace{2em} Reconstruct sparse 3D point cloud

\State \textbf{Bundle Adjustment:}
\Statex \hspace{1em} Refine intrinsic and extrinsic parameters to minimize reprojection error

\State \textbf{Pose Conversion and Dataset Generation:}
\Statex \hspace{1em} Use \texttt{ns-process-data} from Nerfstudio to:
\Statex \hspace{2em} Parse COLMAP outputs
\Statex \hspace{2em} Invert poses: $\mathbf{T}_{wc}^{(i)} = \left( \mathbf{T}_{cw}^{(i)} \right)^{-1}$
\Statex \hspace{2em} Export processed dataset in NeRF-compatible format

\Return Camera-to-world poses $\{ \mathbf{T}_{wc}^{(i)} \}$ and image frames $\{ \mathbf{I}_i \}$
\end{algorithmic}
\end{algorithm}

\subsection{Feature Extraction and Pose Estimation}
After image extraction, COLMAP was employed for feature extraction and pose estimation in~\algoref{alg:pose-estimation}. Feature extraction was performed using COLMAP’s SIFT feature extractor with GPU acceleration enabled to improve computational efficiency. Sequential matching was applied to establish correspondences between frames, ensuring temporal consistency. Unlike exhaustive matching, which checks all possible image pairs, sequential matching assumes an ordered sequence of frames, making it ideal for smooth, linear camera motion while significantly reducing computational complexity~\cite{feng2025wonderverse}.

The Structure-from-Motion (SfM) pipeline was executed using the COLMAP mapper with 64 CPU threads for optimal performance, as the standard COLMAP SfM pipeline runs exclusively on the CPU (COLMAP 3.12.0). This step accounted for the longest processing time in the preprocessing workflow. Although the system supported up to 128 threads, increasing the thread count beyond 64 did not significantly improve processing speed. In fact, the shortest execution time was observed with 64 threads, while 96 and 128 threads yielded similar results. Thus, 64 threads were selected as the preferred configuration. The sparse point cloud was evaluated based on reprojection error, with a target threshold of below 1.0 px~\cite{liu2023comparison}. A bundle adjustment step was applied to optimize intrinsic and extrinsic parameters, refining the camera poses (COLMAP 3.12.0).

We estimated camera poses using COLMAP, which outputs world-to-camera transformation matrices $\mathbf{T}_{cw} \in \mathbb{R}^{4 \times 4}$. Each pose matrix consists of a rotation matrix $\mathbf{R} \in \mathbb{R}^{3 \times 3}$ and a translation vector $\mathbf{t} \in \mathbb{R}^{3}$, such that:

\[
\mathbf{T}_{cw} =
\begin{bmatrix}
\mathbf{R} & \mathbf{t} \\
\mathbf{0} & 1
\end{bmatrix}
\]

To align with NeRF-based pipelines, which expect camera-to-world transformations $\mathbf{T}_{wc}$, we applied the inverse of the COLMAP pose matrices:

\[
\mathbf{T}_{wc} = \mathbf{T}_{cw}^{-1} =
\begin{bmatrix}
\mathbf{R}^{\top} & -\mathbf{R}^{\top} \mathbf{t} \\
\mathbf{0} & 1
\end{bmatrix}
\]

To streamline the preprocessing pipeline, we leveraged the pose conversion functionality in Nerfstudio~\cite{tancik2023nerfstudio}, which parses the COLMAP outputs and performs the necessary inversion to produce camera-to-world transformations suitable for NeRF-based neural rendering. 

This simple preprocessing workflow produces data (poses and images) that convert stationary camera (with rotating object) format into an equivalent moving camera (with stationary object) format that fits into conventional NeRF reconstruction pipelines (camera poses and optimized calibration parameters). 

\subsection{NeRF-Based PCD Reconstruction}
Neural Radiance Fields (NeRFs) have transformed 3D reconstruction by generating high-quality volumetric models directly from 2D images. Rather than relying on traditional mesh-based methods, NeRFs represent a scene as a continuous function over spatial coordinates \((x,y,z)\) and viewing directions. A neural network is trained to predict color and density at every point, effectively capturing the interplay of light within the scene. In our approach, camera poses obtained from COLMAP using a stationary camera are the inputs during training. Once optimized, the network can synthesize novel views and produce precise 3D reconstructions, offering a robust alternative to conventional multi-view stereo techniques. 

We utilized Nerfstudio to train a NeRF model on the preprocessed dataset. The training was conducted using ns-train nerfacto, with normal predictions enabled to enhance surface detail. The model was trained using GPU acceleration. The trained NeRF model was then used to generate a high-resolution 3D point cloud. A Region of Interest (ROI) was defined to remove extraneous data, ensuring the retention of only relevant structures. A 10M-point cloud was exported with outlier removal enabled to improve data quality, as illustrated in \figref{fig:teaser}(right).

\subsection{ Metric Calibration}
Since NeRF inherently produces a normalized coordinate system, it is necessary to rescale the point cloud to recover the original size of the object. In CloudCompare, metric calibration was achieved by scaling the model using a ping-pong ball with a known radius (0.04 m). This calibration step guarantees that the NeRF-based point cloud accurately reflects the object's true physical dimensions, allowing subsequent geometric phenotyping. Following metric calibration, the plant region of interest was segmented from the surrounding data and cleaned using Statistical Outlier Removal (SOR) filtering to minimize noise. This preprocessing step ensured that only the pertinent objects are kept for downstream analysis.

\subsection{Ground Truth Alignment}
We consider the point cloud data constructed using standard NeRF approaches (i.e. moving camera and stationary object) as our ground truth. The rescaled point clouds -- PCD from the stationary camera, and the PCD from the moving camera -- are aligned using the Iterative Closest Point (ICP) algorithm, refining the global registration between the stationary camera NeRF-based reconstruction and the reference measurements. The final aligned point cloud is exported for further accuracy evaluation.

\begin{figure*}[h!] 
    \centering
    \includegraphics[width=\linewidth]{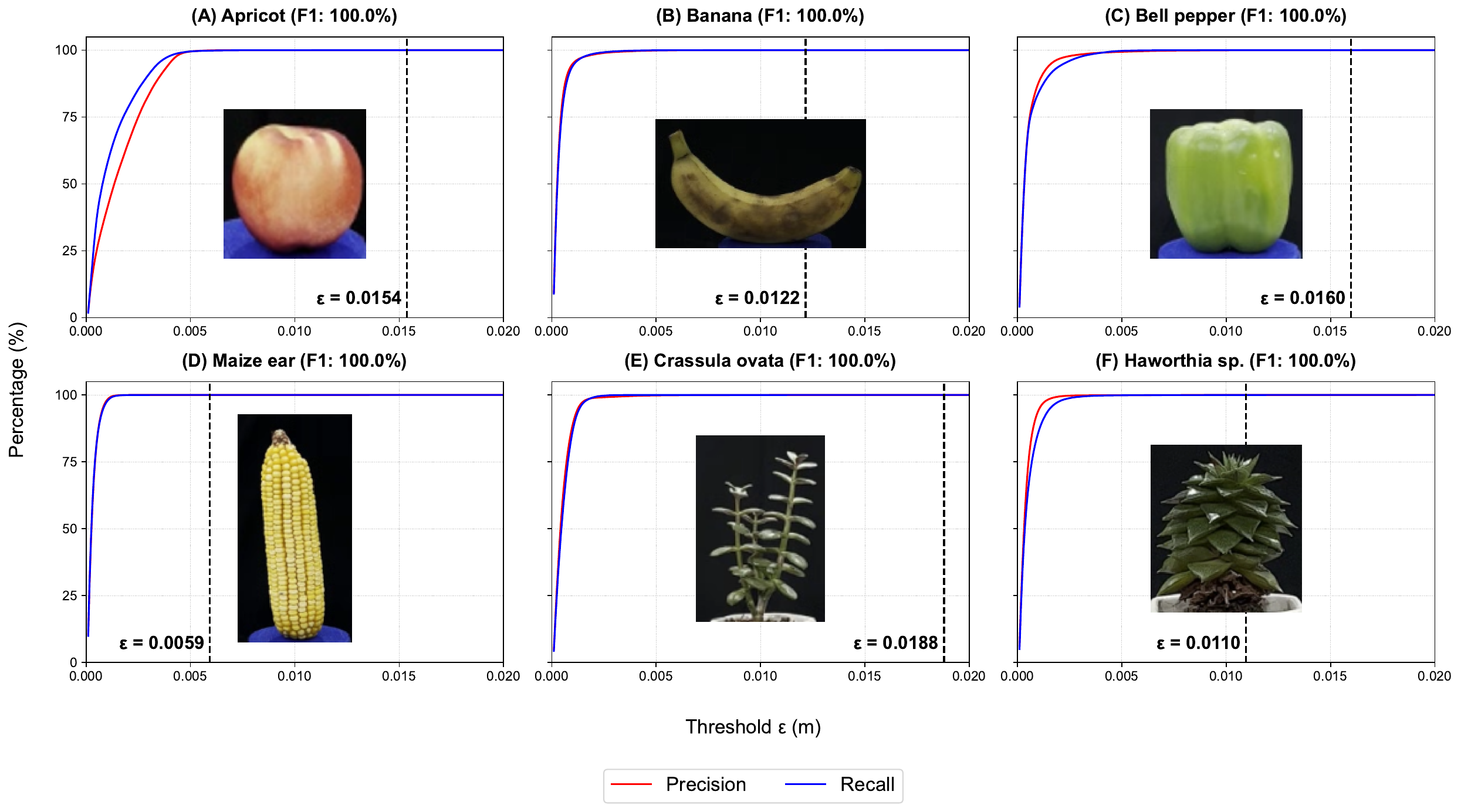}
    \caption{\textbf{Precision-Recall Analysis} for different objects based on varying threshold values. Each plot illustrates the relationship between precision (red) and recall (blue) across different thresholds, with the optimal threshold (\(\epsilon\)) marked by a black dashed line. The F-score for all objects is 100.00, indicating high reconstruction accuracy. The subfigures represent (A) Apricot (\(\epsilon = 0.0110\)), (B) Banana (\(\epsilon = 0.0154\)), (C) Bell pepper (\(\epsilon = 0.0059\)), (D) Maize ear (\(\epsilon = 0.0122\)), (E) \textit{Crassula ovata} (\(\epsilon = 0.0160\)), and (F) \textit{Haworthia sp.} (\(\epsilon = 0.0188\)). This comparison evaluates reconstruction accuracy by analyzing precision and recall behavior at various threshold levels.}
    \label{fig:Precision-Recall}
\end{figure*}

\subsection{Evaluation Metrics}

Following ground truth alignment, the quality of the reconstructed point clouds is quantitatively assessed using precision and recall metrics. In these equations, the set of stationary-camera NeRF reconstructed points is denoted as \(\text{Psc}\) and the set of ground truth points (from standard, moving camera NeRF) as \(\text{Pgt}\). Precision is defined as the ratio of reconstructed points that are within a threshold distance \(\epsilon\) from any ground truth point to the total number of reconstructed points, as shown below:

\[
\text{Precision}(\epsilon) = \frac{\left|\left\{ x \in \text{Psc} \mid \min_{y \in \text{Pgt}} \|x - y\| \leq \epsilon \right\}\right|}{\left|\text{Psc}\right|}
\]

Similarly, recall is the ratio of ground truth points that have a corresponding reconstructed point within the threshold distance to the total number of ground truth points:

\[
\text{Recall}(\epsilon) = \frac{\left|\left\{ y \in \text{Pgt} \mid \min_{x \in \text{Psc}} \|y - x\| \leq \epsilon \right\}\right|}{\left|\text{Pgt}\right|}
\]

These metrics~\cite{arshad2024evaluating} offer a systematic framework for evaluating the spatial accuracy of the NeRF-based reconstruction, ensuring that both the inclusion of relevant details and the exclusion of extraneous data are effectively measured.

\section{Results and Discussion}

\paragraph{Experimental Overview:} The experimental results demonstrate that the proposed stationary camera setup can produce high-quality PCD reconstructions, as illustrated in~\figref{fig:teaser}(right). The reconstructed models closely align with the ground-truth PCD obtained using the rotating camera setup. The primary evaluation metrics included precision-recall analysis (\figref{fig:Precision-Recall}) and computation time comparisons for pose estimation, training, and reconstruction (\tabref{Tab:Compute_time}). The dataset created for this work can be found at \url{https://huggingface.co/datasets/BGLab/SC-NeRF}.

\vfill
\pagebreak

\paragraph{Precision-Recall Analysis:} As shown in~\figref{fig:Precision-Recall}, precision-recall analysis was conducted to evaluate reconstruction accuracy across varying threshold values. The F-score for all tested objects reached 100.00, indicating high reconstruction fidelity. The precision (red) and recall (blue) curves across different thresholds demonstrate that the proposed method effectively captures fine structural details. The optimal threshold ($\epsilon$) values, marked by the black dashed lines, varied slightly among objects, with values ranging from $\epsilon = 0.0059 m$ (Maize ear) for simple structures to $\epsilon = 0.018m$ (\textit{Crassula ovata}) for complex geometries, confirming the model’s adaptability across diverse structures. 

These results validate that the stationary camera setup achieves robust and accurate 3D reconstructions, comparable to those obtained using a rotating camera setup. The combination of high F-scores and precise alignment further supports the effectiveness of the proposed method in capturing detailed object structures with minimal reconstruction error.

\begin{table*}[t!]
    \centering
    \renewcommand{\arraystretch}{1.2}
    \caption{Computation Times for Data Preprocessing, Training, and PCD Reconstruction across different datasets (apricot, banana, corncob, bell pepper, Crassula ovata, and Haworthia sp.). Data preprocessing time is significantly higher for the stationary camera (SC) setup compared to the ground-truth (GT) setup, with the largest difference observed for apricot (27m16.4s vs. 4m41.6s). Training times are generally longer for GT than SC, but the difference remains small, with a maximum gap of 18m14.7s in Haworthia sp.. PCD reconstruction times remain comparable between SC and GT, with differences of only a few minutes across all datasets. The experiments were conducted using the Nova computing cluster with an A100 GPU, 16 CPU cores, and 80 GB of allocated memory.}
    \label{tab:computation_times} 
    
    \begin{tabular}{lcc|cc|cc}
        \toprule
        \textbf{Dataset} & 
        \multicolumn{2}{c|}{\textbf{Data Preprocessing}} & 
        \multicolumn{2}{c|}{\textbf{Training}} & 
        \multicolumn{2}{c}{\textbf{PCD Reconstruction}} \\
        \cmidrule{2-7}
        & GT & SC & GT & SC & GT & SC \\
        \midrule
        Apricot  & 4m41.6s & 27m16.4s & 32m4.1s & 23m14.4s & 5m4.4s  & 4m32.0s \\
        Banana   & 4m34.8s & 11m58.3s & 34m0.2s & 32m8.4s & 4m57.8s & 3m54.9s \\
        Maize Ear     & 6m2.6s & 13m27.1s & 33m4.8s & 33m4.7s & 4m32.9s & 3m32.3s \\
        Bell Pepper  & 5m33.6s & 14m48.9s & 33m8.5s & 33m2.7s & 4m39.43s & 3m37.9s \\
        \textit{Crassula ovata}  & 4m16.5s & 11m45.5s  & 22m33.5s  & 22m36.8s & 5m3.1s  & 5m57.6s \\
        \textit{Haworthia sp.}  & 4m54.0s & 11m45.0s  & 35m3.8s  & 16m58.0s & 5m3.0s  & 5m57.0s \\
        \bottomrule
    \end{tabular}
    \label{Tab:Compute_time}
\end{table*}

\vfill
\pagebreak

\paragraph{Computation Time Analysis:} Computation time analysis revealed that data preprocessing was slower for the stationary camera (SC) setup compared to ground truth (GT), primarily due to the increased time required for pose estimation and feature extraction. As shown in \tabref{Tab:Compute_time}, the largest gap was observed for apricot (27m16.4s for SC vs. 4m41.6s for GT), highlighting the computational burden associated with pose estimation in a stationary camera setting. However, for NeRF training, the differences were relatively small, with SC generally being faster than GT. The largest training time difference was observed in Haworthia sp. (16m58.0s for SC vs. 35m3.8s for GT), showing a difference of approximately 18 minutes in SC.

The PCD reconstruction times between SC and GT were comparable, with differences of only a few minutes across all datasets. The largest discrepancy occurred in Banana (4m57.8s for GT vs. 3m54.9s for SC), showing a 63.8s difference. In contrast, the smallest difference occurred in Apricot (5m4.4s for GT vs. 4m32.0s for SC), with a gap of only 32.4s. These results suggest that although pose estimation is computationally intensive in SC, the overall workflow remains feasible and competitive in terms of training and reconstruction efficiency.

\paragraph{Implications: }The experimental results confirm that NeRF-based PCD reconstruction using a stationary camera setup is both computationally feasible and highly accurate. Although pose estimation remains the primary computational bottleneck, the overall pipeline offers competitive training times and high reconstruction quality, as validated by the precision-recall analysis. These findings demonstrate that high-fidelity 3D reconstructions can be achieved without requiring complex mechanical setups, making the approach well-suited for scalable agricultural imaging applications.

\paragraph{Dataset Availability:}
To support reproducibility, we release the full dataset used in this study as \textbf{SC-NeRF}. It contains all intermediate and final outputs from our NeRF-based reconstruction pipeline for six agriculturally relevant objects: \textit{apricot}, \textit{banana}, \textit{bell pepper}, \textit{maize ear}, \textit{Crassula ovata}, and \textit{Haworthia sp.}. Each object was recorded under two settings:
\textbf{SC}: Stationary camera, rotating object (our method). \textbf{GT}: Moving camera, stationary object (used as ground truth). The dataset is organized as follows:

\vspace{0.5em}
\noindent\textbf{raw/}: Includes 4K videos (.MOV) and extracted frames.\\
\textbf{pre/}: Contains COLMAP camera poses and sparse reconstructions, formatted via \texttt{ns-process-data}.\\
\textbf{train/}: Holds trained NeRF model checkpoints using the \texttt{nerfacto} trainer.\\
\textbf{pcd/}: Final aligned and scaled 10M-point clouds, filtered via outlier removal.

\vspace{0.5em}
\noindent Each file is named according to the object and capture mode (e.g., \texttt{banana\_sc}, \texttt{apricot\_gt}). The dataset can support training, benchmarking, or extending NeRF-based 3D reconstruction pipelines in agriculture. The dataset can be found at \url{https://huggingface.co/datasets/BGLab/SC-NeRF}.

\section{Conclusions}
This paper introduces a NeRF-based point cloud (PCD) reconstruction framework explicitly designed for indoor high-throughput plant phenotyping environments. Traditional NeRF methods require cameras to move around stationary objects, a process incompatible with automated indoor phenotyping facilities that routinely employ stationary cameras alongside rotating pedestals or conveyors. To overcome this limitation, we develop a stationary-camera-based NeRF approach that simulates camera motion through a straightforward pose transformation after COLMAP-based pose estimation, facilitating standard NeRF training. Our experimental validation demonstrated that this method achieves high-fidelity 3D reconstructions, yielding very high precision-recall F-scores across various agriculturally relevant objects. Despite the computational time associated with pose estimation in stationary setups, our framework showed competitive overall reconstruction times, highlighting its practical feasibility for integration into automated phenotyping workflows.


This work has direct relevance for plant science research, breeding programs, and agricultural production, particularly when employing expensive or fragile imaging instruments such as hyperspectral cameras. By eliminating the need for complex camera rigs and costly imaging hardware, our method simplifies high-throughput phenotyping infrastructure, reducing both operational complexity and expenses. Future research directions will involve optimizing pose estimation processes to reduce computational demand, exploring the method's adaptability to a broader range of plant species and phenotyping scenarios, and integrating multimodal data such as hyperspectral bands with RGB imaging.

\clearpage
{
    \small
    \bibliographystyle{ieeenat_fullname}
    \bibliography{main}

\begin{thebibliography}{34}
\providecommand{\natexlab}[1]{#1}
\providecommand{\url}[1]{\texttt{#1}}
\expandafter\ifx\csname urlstyle\endcsname\relax
  \providecommand{\doi}[1]{doi: #1}\else
  \providecommand{\doi}{doi: \begingroup \urlstyle{rm}\Url}\fi

\bibitem[And{\'u}jar et~al.(2018)And{\'u}jar, Calle, Fern{\'a}ndez-Quintanilla, Ribeiro, and Dorado]{andujar2018three}
Dionisio And{\'u}jar, Mikel Calle, C{\'e}sar Fern{\'a}ndez-Quintanilla, {\'A}ngela Ribeiro, and Jos{\'e} Dorado.
\newblock Three-dimensional modeling of weed plants using low-cost photogrammetry.
\newblock \emph{Sensors}, 18\penalty0 (4):\penalty0 1077, 2018.

\bibitem[Arshad et~al.(2024)Arshad, Jubery, Afful, Jignasu, Balu, Ganapathysubramanian, Sarkar, and Krishnamurthy]{arshad2024evaluating}
Muhammad~Arbab Arshad, Talukder Jubery, James Afful, Anushrut Jignasu, Aditya Balu, Baskar Ganapathysubramanian, Soumik Sarkar, and Adarsh Krishnamurthy.
\newblock Evaluating neural radiance fields for 3d plant geometry reconstruction in field conditions.
\newblock \emph{Plant Phenomics}, 6:\penalty0 0235, 2024.

\bibitem[Blancon et~al.(2024)Blancon, Buet, Dubreuil, Tixier, Baret, and Praud]{blancon2024maize}
Justin Blancon, Cl{\'e}ment Buet, Pierre Dubreuil, Marie-H{\'e}l{\`e}ne Tixier, Fr{\'e}d{\'e}ric Baret, and S{\'e}bastien Praud.
\newblock Maize green leaf area index dynamics: genetic basis of a new secondary trait for grain yield in optimal and drought conditions.
\newblock \emph{Theoretical and Applied Genetics}, 137\penalty0 (3):\penalty0 68, 2024.

\bibitem[Chen et~al.(2024)Chen, Narayanan, Ottou, Missaoui, Muriki, Pradalier, and Chen]{chen2024hyperspectral}
Gerry Chen, Sunil~Kumar Narayanan, Thomas~Gautier Ottou, Benjamin Missaoui, Harsh Muriki, C{\'e}dric Pradalier, and Yongsheng Chen.
\newblock Hyperspectral neural radiance fields.
\newblock \emph{arXiv preprint arXiv:2403.14839}, 2024.

\bibitem[Chen et~al.(2019)Chen, Han, Xu, and Su]{chen2019point}
Rui Chen, Songfang Han, Jing Xu, and Hao Su.
\newblock Point-based multi-view stereo network.
\newblock In \emph{Proceedings of the IEEE/CVF international conference on computer vision}, pages 1538--1547, 2019.

\bibitem[Cuevas-Velasquez et~al.(2020)Cuevas-Velasquez, Gallego, and Fisher]{cuevas2020segmentation}
Hanz Cuevas-Velasquez, Antonio-Javier Gallego, and Robert~B Fisher.
\newblock Segmentation and {3D} reconstruction of rose plants from stereoscopic images.
\newblock \emph{Computers and electronics in agriculture}, 171:\penalty0 105296, 2020.

\bibitem[Delattre et~al.(2023)Delattre, Dirnfeld, Nguyen, Scarano, Jones, Miraldo, and Learned-Miller]{delattre2023robust}
Fabien Delattre, David Dirnfeld, Phat Nguyen, Stephen~K Scarano, Michael~J Jones, Pedro Miraldo, and Erik Learned-Miller.
\newblock Robust frame-to-frame camera rotation estimation in crowded scenes.
\newblock In \emph{Proceedings of the IEEE/CVF International Conference on Computer Vision}, pages 9752--9762, 2023.

\bibitem[Eltner and Sofia(2020)]{eltner2020structure}
Anette Eltner and Giulia Sofia.
\newblock Structure from motion photogrammetric technique.
\newblock In \emph{Developments in Earth surface processes}, pages 1--24. Elsevier, 2020.

\bibitem[Feng et~al.(2025)Feng, Zuo, Pan, Hui, Shao, Dou, Xie, and Liu]{feng2025wonderverse}
Hao Feng, Zhi Zuo, Jia-hui Pan, Ka-hei Hui, Yi-hua Shao, Qi Dou, Wei Xie, and Zheng-zhe Liu.
\newblock Wonderverse: Extendable 3d scene generation with video generative models.
\newblock \emph{arXiv preprint arXiv:2503.09160}, 2025.

\bibitem[Feng et~al.(2023)Feng, Saadati, Jubery, Jignasu, Balu, Li, Attigala, Schnable, Sarkar, Ganapathysubramanian, et~al.]{feng20233d}
Jiale Feng, Mojdeh Saadati, Talukder Jubery, Anushrut Jignasu, Aditya Balu, Yawei Li, Lakshmi Attigala, Patrick~S Schnable, Soumik Sarkar, Baskar Ganapathysubramanian, et~al.
\newblock {3D} reconstruction of plants using probabilistic voxel carving.
\newblock \emph{Computers and Electronics in Agriculture}, 213:\penalty0 108248, 2023.

\bibitem[Gao et~al.(2023)Gao, Su, Liang, Yue, Yang, and Fu]{gao2023mc}
Yu Gao, Lutong Su, Hao Liang, Yufeng Yue, Yi Yang, and Mengyin Fu.
\newblock Mc-nerf: Multi-camera neural radiance fields for multi-camera image acquisition systems.
\newblock \emph{arXiv preprint arXiv:2309.07846}, 2023.

\bibitem[Grys et~al.(2017)Grys, Lo, Sahin, Kraus, Morris, Boone, and Andrews]{grys2017machine}
Ben~T Grys, Dara~S Lo, Nil Sahin, Oren~Z Kraus, Quaid Morris, Charles Boone, and Brenda~J Andrews.
\newblock Machine learning and computer vision approaches for phenotypic profiling.
\newblock \emph{Journal of Cell Biology}, 216\penalty0 (1):\penalty0 65--71, 2017.

\bibitem[Gupta et~al.(2024)Gupta, Pagani, Zamboni, and Singh]{gupta2024ai}
Deependra~Kumar Gupta, Anselmo Pagani, Paolo Zamboni, and Ajay~Kumar Singh.
\newblock Ai-powered revolution in plant sciences: advancements, applications, and challenges for sustainable agriculture and food security.
\newblock \emph{Exploration of Foods and Foodomics}, 2\penalty0 (5):\penalty0 443--459, 2024.

\bibitem[Hirano et~al.(2009)Hirano, Funayama, and Maekawa]{hirano20093d}
Daisuke Hirano, Yusuke Funayama, and Takashi Maekawa.
\newblock 3d shape reconstruction from 2d images.
\newblock \emph{Computer-Aided Design and Applications}, 6\penalty0 (5):\penalty0 701--710, 2009.

\bibitem[Hu et~al.(2024)Hu, Ying, Pan, Kang, and Chen]{hu2024high}
Kewei Hu, Wei Ying, Yaoqiang Pan, Hanwen Kang, and Chao Chen.
\newblock High-fidelity 3d reconstruction of plants using neural radiance fields.
\newblock \emph{Computers and Electronics in Agriculture}, 220:\penalty0 108848, 2024.

\bibitem[Hu et~al.(2020)Hu, Zhang, Jiang, Huang, Liu, Xiong, Yang, and Chen]{hu2020nondestructive}
Weijuan Hu, Can Zhang, Yuqiang Jiang, Chenglong Huang, Qian Liu, Lizhong Xiong, Wanneng Yang, and Fan Chen.
\newblock Nondestructive 3d image analysis pipeline to extract rice grain traits using x-ray computed tomography.
\newblock \emph{Plant Phenomics}, 2020.

\bibitem[Jignasu et~al.(2023)Jignasu, Herron, Jubery, Afful, Balu, Ganapathysubramanian, Sarkar, and Krishnamurthy]{jignasu2023plant}
Anushrut Jignasu, Ethan Herron, Talukder~Zaki Jubery, James Afful, Aditya Balu, Baskar Ganapathysubramanian, Soumik Sarkar, and Adarsh Krishnamurthy.
\newblock Plant geometry reconstruction from field data using neural radiance fields.
\newblock In \emph{2nd AAAI Workshop on AI for Agriculture and Food Systems}, 2023.

\bibitem[Kusmec et~al.(2018)Kusmec, de~Leon, and Schnable]{kusmec2018harnessing}
Aaron Kusmec, Natalia de Leon, and Patrick~S Schnable.
\newblock Harnessing phenotypic plasticity to improve maize yields.
\newblock \emph{Frontiers in Plant Science}, 9:\penalty0 1377, 2018.

\bibitem[Lei et~al.(2023)Lei, Liu, Xie, Fang, Wang, Luo, Li, Yu, and Qiu]{lei20233d}
Shuhan Lei, Li Liu, Yu Xie, Ying Fang, Chuangxia Wang, Ninghao Luo, Ruitao Li, Donghai Yu, and Zixuan Qiu.
\newblock 3d visualization technology for rubber tree forests based on a terrestrial photogrammetry system.
\newblock \emph{Frontiers in Forests and Global Change}, 6:\penalty0 1206450, 2023.

\bibitem[Levy et~al.(2023)Levy, Matthews, Sela, Wetzstein, and Lagun]{levy2023melon}
Axel Levy, Mark Matthews, Matan Sela, Gordon Wetzstein, and Dmitry Lagun.
\newblock Melon: Nerf with unposed images using equivalence class estimation.
\newblock \emph{arXiv preprint arXiv:2303.08096}, 2, 2023.

\bibitem[Liu et~al.(2023)Liu, Bonelli, Pietrzyk, and Bucksch]{liu2023comparison}
Suxing Liu, Wesley~Paul Bonelli, Peter Pietrzyk, and Alexander Bucksch.
\newblock Comparison of open-source three-dimensional reconstruction pipelines for maize-root phenotyping.
\newblock \emph{The Plant Phenome Journal}, 6\penalty0 (1):\penalty0 e20068, 2023.

\bibitem[Lu(2023)]{lu2023bird}
Guoyu Lu.
\newblock Bird-view {3D} reconstruction for crops with repeated textures.
\newblock In \emph{IEEE/RSJ International Conference on Intelligent Robots and Systems (IROS)}, pages 4263--4270. IEEE, 2023.

\bibitem[Medic et~al.(2023)Medic, B{\"o}mer, and Paulus]{medic2023challenges}
Tomislav Medic, Jonas B{\"o}mer, and Stefan Paulus.
\newblock Challenges and recommendations for 3d plant phenotyping in agriculture using terrestrial lasers scanners.
\newblock \emph{ISPRS Annals of the Photogrammetry, Remote Sensing and Spatial Information Sciences}, 10:\penalty0 1007--1014, 2023.

\bibitem[Mildenhall et~al.(2021)Mildenhall, Srinivasan, Tancik, Barron, Ramamoorthi, and Ng]{mildenhall2021nerf}
Ben Mildenhall, Pratul~P Srinivasan, Matthew Tancik, Jonathan~T Barron, Ravi Ramamoorthi, and Ren Ng.
\newblock {NeRF}: Representing scenes as neural radiance fields for view synthesis.
\newblock \emph{Communications of the ACM}, 65\penalty0 (1):\penalty0 99--106, 2021.

\bibitem[Paturkar et~al.(2021)Paturkar, Gupta, and Bailey]{paturkar2021effect}
Abhipray Paturkar, Gourab~Sen Gupta, and Donald Bailey.
\newblock Effect on quality of {3D} model of plant with change in number and resolution of images used: {An} investigation.
\newblock In \emph{Advances in Signal and Data Processing: Select Proceedings of ICSDP 2019}, pages 377--388. Springer, 2021.

\bibitem[Rudnev et~al.(2023)Rudnev, Elgharib, Theobalt, and Golyanik]{rudnev2023eventnerf}
Viktor Rudnev, Mohamed Elgharib, Christian Theobalt, and Vladislav Golyanik.
\newblock Eventnerf: Neural radiance fields from a single colour event camera.
\newblock In \emph{Proceedings of the IEEE/CVF Conference on Computer Vision and Pattern Recognition}, pages 4992--5002, 2023.

\bibitem[Sarkar et~al.(2023)Sarkar, Ganapathysubramanian, Singh, Fotouhi, Kar, Nagasubramanian, Chowdhary, Das, Kantor, Krishnamurthy, Merchant, and Singh]{sarkar2023cyber}
Soumik Sarkar, Baskar Ganapathysubramanian, Arti Singh, Fateme Fotouhi, Soumyashree Kar, Koushik Nagasubramanian, Girish Chowdhary, Sajal~K Das, George Kantor, Adarsh Krishnamurthy, Nirav Merchant, and Asheesh~K. Singh.
\newblock Cyber-agricultural systems for crop breeding and sustainable production.
\newblock \emph{Trends in Plant Science}, 2023.

\bibitem[Tancik et~al.(2023)Tancik, Weber, Ng, Li, Yi, Wang, Kristoffersen, Austin, Salahi, Ahuja, et~al.]{tancik2023nerfstudio}
Matthew Tancik, Ethan Weber, Evonne Ng, Ruilong Li, Brent Yi, Terrance Wang, Alexander Kristoffersen, Jake Austin, Kamyar Salahi, Abhik Ahuja, et~al.
\newblock {NeRFStudio}: A modular framework for neural radiance field development.
\newblock In \emph{ACM SIGGRAPH Conference Proceedings}, pages 1--12, 2023.

\bibitem[Tang et~al.(2022)Tang, Wang, Wang, Guo, and Zhang]{tang2022benefits}
Xiaoying Tang, Mengjun Wang, Qian Wang, Jingjing Guo, and Jingxiao Zhang.
\newblock Benefits of terrestrial laser scanning for construction qa/qc: a time and cost analysis.
\newblock \emph{Journal of Management in Engineering}, 38\penalty0 (2):\penalty0 1--10, 2022.

\bibitem[Tucker et~al.(2020)Tucker, Dohleman, Grapov, Flagel, Yang, Wegener, Kosola, Swarup, Rapp, Bedair, et~al.]{tucker2020evaluating}
Sarah~L Tucker, Frank~G Dohleman, Dmitry Grapov, Lex Flagel, Sean Yang, Kimberly~M Wegener, Kevin Kosola, Shilpa Swarup, Ryan~A Rapp, Mohamed Bedair, et~al.
\newblock Evaluating maize phenotypic variance, heritability, and yield relationships at multiple biological scales across agronomically relevant environments.
\newblock \emph{Plant, cell \& environment}, 43\penalty0 (4):\penalty0 880--902, 2020.

\bibitem[Westhues et~al.(2021)Westhues, Mahone, da~Silva, Thorwarth, Schmidt, Richter, Simianer, and Beissinger]{westhues2021prediction}
Cathy~C Westhues, Gregory~S Mahone, Sofia da Silva, Patrick Thorwarth, Malthe Schmidt, Jan-Christoph Richter, Henner Simianer, and Timothy~M Beissinger.
\newblock Prediction of maize phenotypic traits with genomic and environmental predictors using gradient boosting frameworks.
\newblock \emph{Frontiers in plant science}, 12:\penalty0 699589, 2021.

\bibitem[Wu et~al.(2025)Wu, Hu, Tian, Huang, Yang, Li, and Xu]{wu163d}
Sixiao Wu, Changhao Hu, Boyuan Tian, Yuan Huang, Shuo Yang, Shanjun Li, and Shengyong Xu.
\newblock A {3D} reconstruction platform for complex plants using {OB-NeRF}.
\newblock \emph{Frontiers in Plant Science}, 16:\penalty0 1449626, 2025.

\bibitem[Young et~al.(2023)Young, Jubery, Carley, Carroll, Sarkar, Singh, Singh, and Ganapathysubramanian]{young2023canopy}
Therin~J. Young, Talukder~Z. Jubery, Carley~N. Carley, M. Carroll, Soumik Sarkar, Asheesh~K. Singh, Arti Singh, and Baskar Ganapathysubramanian.
\newblock “canopy fingerprints” for characterizing three-dimensional point cloud data of soybean canopies.
\newblock \emph{Frontiers in Plant Science}, 14:\penalty0 1141153, 2023.

\bibitem[Zhang and Wong(2008)]{zhang2008self}
Hui Zhang and Kwan-Yee~K Wong.
\newblock Self-calibration of turntable sequences from silhouettes.
\newblock \emph{IEEE transactions on pattern analysis and machine intelligence}, 31\penalty0 (1):\penalty0 5--14, 2008.

\end{thebibliography}
}


%

\end{document}